\documentclass[10pt,journal,compsoc,twoside]{IEEEtran}
\ifCLASSOPTIONcompsoc
  \usepackage[nocompress]{cite}
\else
  \usepackage{cite}
\fi
\usepackage{array}
\usepackage{graphicx}
\graphicspath{{figures/}}
\usepackage{amsmath}
\usepackage{dblfloatfix}
\usepackage{url}
\usepackage{orcidlink}
\usepackage{comment}

\title{EEGminer: Discovering Interpretable Features of Brain Activity with Learnable Filters}

\author{Siegfried~Ludwig~\orcidlink{0000-0001-9693-9482},
        Stylianos~Bakas~\orcidlink{0000-0003-1054-0169},
        Dimitrios~A.~Adamos~\orcidlink{0000-0001-6700-1057},\\
        Nikolaos~Laskaris~\orcidlink{0000-0002-1960-394X},
        Yannis~Panagakis~\orcidlink{0000-0003-0153-5210},
        Stefanos~Zafeiriou~\orcidlink{0000-0002-5222-1740},
\IEEEcompsocitemizethanks{
    \IEEEcompsocthanksitem S.~Ludwig and S.~Bakas (equal contribution), D.~Adamos and S.~Zafeiriou are with the Department
    of Computing, Imperial College London, London SW7 2RH, U.K.\protect\\
    E-mail: siegfried.ludwig20@imperial.ac.uk
    \IEEEcompsocthanksitem S. Bakas and N.~Laskaris are with the School of Informatics, Aristotle University of Thessaloniki, Thessaloniki 54124, Greece.\protect\\
    E-mail: simpakas@csd.auth.gr
    \IEEEcompsocthanksitem D.~Adamos is with the School of Music Studies, Aristotle University of Thessaloniki, Thessaloniki 54124, Greece.
    \IEEEcompsocthanksitem Y.~Panagakis is with the Department of Informatics and Telecommunications, National and Kapodistrian University of Athens, Athens 15784, Greece.
    \IEEEcompsocthanksitem All authors are with Cogitat Ltd., London, U.K.
    }%
}

\IEEEtitleabstractindextext{%
\begin{abstract}
Patterns of brain activity are associated with different brain processes and can be used to identify different brain states and make behavioral predictions. However, the relevant features are not readily apparent and accessible. To mine informative latent representations from multichannel recordings of ongoing EEG activity, we propose a novel differentiable decoding pipeline consisting of learnable filters and a pre-determined feature extraction module. Specifically, we introduce filters parameterized by generalized Gaussian functions
that offer a smooth derivative for stable end-to-end model training and allow for learning interpretable features. For the feature module, we use signal magnitude and functional connectivity estimates. We demonstrate the utility of our model towards emotion recognition from EEG signals on the SEED dataset, as well as on a new EEG dataset of unprecedented size (i.e., 761 subjects), where we identify consistent trends of music perception and related individual differences. The discovered features align with previous neuroscience studies and offer new insights, such as marked differences in the functional connectivity profile between left and right temporal areas during music listening. This agrees with the respective specialisation of the temporal lobes regarding music perception proposed in the literature.
\end{abstract}

}

\begin{document}

\maketitle

\IEEEdisplaynontitleabstractindextext

\IEEEpeerreviewmaketitle

\IEEEraisesectionheading{\section{Introduction}\label{sec:introduction}}

%[Brain decoding]
\IEEEPARstart{H}{uman} behavior arises from the interactions between many different brain regions \cite{friston2011functional}\cite{bassett2017network}. Hence, revealing and decoding functional brain connectivity enables making predictions of behavioral states \cite{shirer2012decoding}, identifying subjects and predicting fluid intelligence \cite{finn2015functional}, as well as facilitates predicting sustained attention \cite{rosenberg2016neuromarker} and emotional responses to music \cite{shahabi2016toward}\cite{bakas2021estimate}, to mention but a few examples of important applications. However, the underlying brain connectivity is not readily accessible from EEG recordings, which simply record electric potential differences at multiple locations on the skull.

%[Finding features]
Several conventional measures have been proposed for estimating underlying connectivity from EEG recordings \cite{kida2016multi}, such as signal correlations and phase locking values \cite{lachaux1999measuring}. Tuning these measures, however, involves extensive manual feature engineering. Indeed, in practice, the search space of informative features explodes exponentially when one considers coordinated brain activity and employs functional interactions between distinct brain areas. Therefore, manual feature engineering is a suboptimal process. In contrast, deep neural networks \cite{lecun2015deep} enable learning features from large amounts of annotated training data. Concretely, such deep models involve trainable parameters (usually arranged into a series of processing layers), the values of which are estimated using backpropagation of gradients \cite{lecun1989backpropagation} from the training data. However, features extracted by deep models often lack in interpretability, which can present significant detriments in real-world applications \cite{rudin2019stop}. Besides interpretability, off-the-shelf deep learning models require a significant amount of annotated data, hindering their applicability in data-scarce scenarios such as EEG analysis.

% Related work summary
The aforementioned limitations of deep neural networks regarding interpretability and the need for large training datasets can be reduced by introducing appropriate inductive biases. The most common approach in the EEG domain is to use convolutional neural networks (CNN) \cite{craik2019deep}\cite{roy2019deep}. A prominent example is EEGNet, which takes cues from conventional EEG signal processing techniques to create a compact and relatively interpretable model \cite{lawhern2018eegnet}. There are very limited constraints on the model however, such as fully trainable temporal filter functions. Convolutional filters have been constrained by parameterized sinc functions in the audio domain \cite{ravanelli2018speaker}. This constrains the model but could reduce trainability, especially in the low signal-to-noise ratio regime of the EEG domain. Further, CNNs are usually thought of as processing spectral-temporal features. An attempt to introduce constraints towards using EEG functional connectivity features has been undertaken with SyncNet \cite{li2017targeting}, which however does not make these constraints fully explicit and thereby limits its interpretability.

\begin{figure*}[!t]
    \centering
    \includegraphics[width=\textwidth]{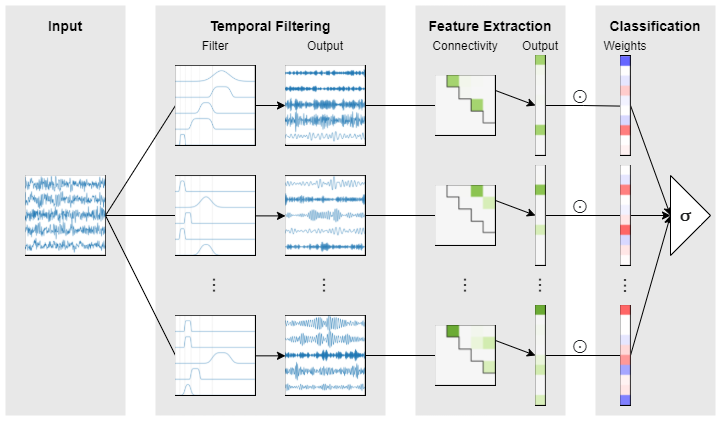}
    \caption{Visualization of the EEGminer architecture. Independently trainable filters are applied on each input channel (five signal channels are given here over four seconds). As an example of a differentiable feature module used in this paper, the filtered signals are then correlated with each other to get connectivity matrices. As they are symmetric, the upper triangular is extracted as a feature vector. Multiple filters per input channel result in multiple independent feature maps (shown here are three connectivity feature maps). The feature vectors resulting from all feature maps are concatenated, standardized (not shown) and used by the linear classifier for logistic regression.}
    \label{fig:EEGminer}
\end{figure*}

% Our proposed method
Distinct from the above models, we propose to implement an end-to-end differentiable EEG decoding pipeline (coined as EEGminer), making use of explicit neuroscientific measures of brain activity, such as signal correlations. The type of feature itself is therefore untrainable, but by using a differentiable implementation, the frequencies over which the connectivity is computed can be learned. The proposed model consists of learnable temporal filters parameterized by generalized Gaussians, a differentiable feature module and a linear layer for classification (Fig. \ref{fig:EEGminer}). Our approach can easily incorporate multiple types of features (local activity, covariation, phase synchrony, etc.) and keeps close relation with the topological information about the sensors, which in turn guarantees the direct interpretation of the trained model. Our approach can be interpreted both from the perspectives of conventional neuroscience on the one hand and deep learning on the other. The steps that make up an EEG pipeline (e.g. signal filtering, connectivity measure, classification) are analogous to consecutive layers in a deep neural network. Conversely, viewed from a deep learning perspective, our proposal for a differentiable EEG pipeline can be seen as introducing strong inductive biases into an end-to-end trainable model. By choosing an appropriate model architecture and training procedure, the interpretability of learned features can be made explicit while maintaining competitive decoding accuracy.

% Generalised Gaussian filters
Learnable Gaussian filters have been used before as part of deep learning models, both as a trainable filterbank \cite{seki2017deep} as well as in their time-domain representation as Morlet wavelets to parameterize a convolutional filter \cite{zhao2019learning}. In this work, we propose to design a more general filter function, based on the generalized Gaussian, which introduces a shape parameter. During training, the filter shape can adapt from a Gaussian (i.e. Morlet wavelet) with good training behavior to a rectangular filter (i.e. sinc) with good frequency selectivity.

%[Datasets, tasks and baseline models]
To demonstrate the generalizability of our approach, we apply it on two datasets and three distinct classification tasks, in all cases searching for subject-independent features. The chosen tasks put a focus on spontaneous brain activity that does not exhibit event-locked responses. The main dataset used was collected in collaboration with the Science Museum London, providing recordings from an unprecedented number of participants during passive music listening. We train our model to discriminate between resting state and music listening and to differentiate female and male subjects based on their brainwaves during music listening. Generalizing to a different setting, we use the SEED dataset to classify positive and negative emotions elicited by watching emotional film clips \cite{duan2013differential}\cite{zheng2015investigating}. For all three tasks we compare model classification performance to conventional and deep learning approaches and extract the learned magnitude and functional connectivity features and the importance attributed to them by the classifier.

In summary the contributions of this paper are as follows:
\begin{itemize}
    \item We introduce generalised Gaussian filters that allow for end-to-end learning of frequency bands of interest
    \item We propose a modular deep learning architecture for decoding EEG signals with explicitly interpretable features, without sacrificing accuracy
    \item We provide novel insights into brain states underlying music listening and emotional responses
%    \item Is the dataset one of the main contributions of the paper?
\end{itemize}

\begin{figure*}[!t]
    \centering
    \includegraphics[width=\textwidth]{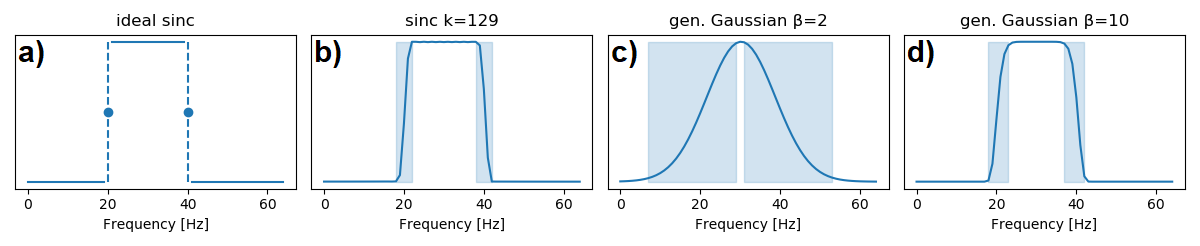}
    \caption{Frequency domain representations of a) ideal sinc bandpass filter, b) sinc bandpass filter truncated in time domain to kernel size=129 with Hamming window, c) generalised Gaussian filter with shape parameter $\beta$=2, corresponding to a Morlet wavelet, and d) generalised Gaussian with $\beta$=10. Shaded areas denote regions of the filter functions that provide gradients, i.e. non-zero derivatives.}
    \label{fig:filter_gradients}
\end{figure*}

The remainder of the paper is organized as follows: Section \ref{sec:related_work} describes related work in the literature and section \ref{sec:methods} outlines the proposed method. Section \ref{sec:evaluation} describes the baselines and datasets employed in this study and presents results on performance benchmarks as well as interpretations of the learned features and their statistical validation. The last section summarizes and concludes our work.

\section{Related Work}
\label{sec:related_work}

% ShallowConvNet
Two widely cited convolutional neural networks developed for EEG decoding are ShallowConvNet and DeepConvNet \cite{schirrmeister2017deep}. ShallowConvNet employs a single temporal filtering layer with trainable convolution kernels shared across electrode channels, followed by a spatial filter, temporal pooling and linear classifier. The composition of temporal and spatial filters is inspired by the filter bank common spatial pattern algorithm \cite{ang2012filter}. Using similar operations, DeepConvNet targets more complex feature extraction with additional network layers. Both models have a large number of trainable parameters due to missing constraints and are relatively difficult to interpret as a consequence.

% EEGNet
EEGNet is another important convolutional decoding model developed for the EEG domain \cite{lawhern2018eegnet}. The EEGNet architecture is comprised of trainable convolutional temporal filters very similar to ShallowConvNet. This is followed by spatial filters, which create projections from signal channels filtered in the same frequencies. Constraining the spatial filter to work within a single temporal filter channel makes the model more parameter efficient and facilitates learning from relatively small datasets of EEG recordings. The model applies a further convolution layer for higher-level feature extraction and a linear classifier. Being similar in its architecture to ShallowConvNet, the interpretability is limited.

% SincNet
In order to reduce model complexity, parameterized sinc bandpass filters can be employed instead of arbitrary convolutional filters (i.e., SincNet \cite{ravanelli2018speaker} and related architectures \cite{borra2020interpretable}\cite{bria2021sinc}). The kernel of a convolutional sinc bandpass filter is constructed as the difference between two sinc lowpass filters as
\begin{equation}
    k(n) = 2 f_2 sinc(2 \pi f_2 n) - 2 f_1 sinc(2 \pi f_1 n),
\end{equation}
where the sinc function is defined as $sinc(x)=sin(x)/x$. The trainable parameters are here only the low and high cutoff frequencies $f_1$ and $f_2$ as opposed to the full filter function. Limiting the temporal filters to sinc functions improves interpretability, since each filter is defined by a specific frequency band, while still allowing for learnable filters. However, an important difference between audio and EEG signals is the significantly lower signal-to-noise ratio in the latter, which is expected to make training of the bandpass filters more difficult. Looking at sinc bandpass filters in the frequency domain, a weakness with regard to its use in deep learning is apparent: the ideal sinc bandpass filter has a rectangular frequency response, which has a flat pass-band and essentially no transition band and therefore a zero derivative at essentially every point, meaning it cannot give useful gradients when training a deep learning model (see Fig. \ref{fig:filter_gradients}). The convolutional sinc filter still works, since it has to be truncated in the time domain, which introduces a transition band in the frequency domain and therefore some useful gradients, but this leaves room for improvement.

% SyncNet
Applying a similar constraint on the first-level filters, recent work on inductive biases for the EEG domain has been done with SyncNet, a simple deep learning model targeting EEG synchrony with claims towards interpretability \cite{li2017targeting}. The model is based on a trainable convolutional layer parameterized as the real part of a complex Morlet wavelet. The time-domain filter is defined as
\begin{equation}
    k(n) = b \; cos(\omega n + \phi) \, exp(- \beta n^2),
\end{equation}
where $b$ is the amplitude of the wavelet, $\omega$ is the center frequency, $\phi$ is the phase-shift and $\beta$ controls the frequency-time precision tradeoff. An approximation of cross-spectral density is then achieved by a linear combination of the filtered input channels. This linear combination is identical in its nature to the spatial projection used in EEGNet and is thereby limited in its approximation of synchrony. Furthermore, it insufficiently constrains the feature space towards an approximation of synchrony.

% SPDNet
Convolutional neural networks are commonly thought to work with spectral features. Recently there has been growing interest in adapting deep learning architectures to work on symmetric positive definite (SPD) matrices, such as spatial covariance matrices of EEG signals. SPDNet was developed as a simple deep learning model to be applied on SPD matrices \cite{huang2017riemannian} and has been used for EEG decoding \cite{hajinoroozi2017prediction}. The focus with this model is however on the classification of SPD matrices, not on creating the specific SPD matrix from input signals in the first place. In the case of EEG decoding, usually a covariance matrix derived from wide-band signals is used. The interpretation of trained features is difficult.

% Desirable model characteristics
Considering the limitations of the aforementioned models, a well trainable and directly interpretable model should have the following characteristics:
\begin{itemize}
    \item The types of temporal filters should be explicitly defined and have good gradient support on all relevant frequencies
    \item The feature extraction should be a close approximation or direct computation, and well constrained to the specific feature
    \item The training procedure should enable direct interpretation of classifier weights
\end{itemize}

\section{Proposed Method}
\label{sec:methods}

The proposed model consists of trainable temporal filters with specific constraints, followed by a pre-defined differentiable feature module (band magnitude, functional connectivity) and leading to a linear classifier (see Fig. \ref{fig:EEGminer}). Each of these components will be presented in the following subsections, accompanied by our strategy for feature interpretation and the training procedure.

\subsection{Temporal Filters}

As mentioned in section \ref{sec:related_work}, a prominent example of using parameterized filter functions in a deep learning model are sinc bandpass filters \cite{ravanelli2018speaker}. Sinc filters however have a very narrow transition band, which means the derivative of the filter function with regard to most frequencies is zero (see Fig. \ref{fig:filter_gradients}). The filter will therefore get limited gradients during training. In order to improve trainability, we propose generalized Gaussian filters in the frequency domain, which have a wide transition band for good gradient support.

Filtering in the frequency domain allows for full control of the filter function \cite{widmann2015digital} and has previously been used in a deep learning model \cite{brosch2015efficient}, confirming that it does not impede differentiability. The generalised Gaussian is given by
\begin{equation}
    G(x)=\frac{\beta}{2\alpha\Gamma(1/\beta)} e^{-(|x-\mu|/\alpha)^{\beta}},
\end{equation}
where $\Gamma(n)=(n-1)!$ is the gamma function, $\mu$ is the center frequency, $\alpha$ is the scale and $\beta$ is the shape parameter. We fix the gain of all filters at one and accordingly drop the normalization factor of the generalised Gaussian.

The bandwidth of a filter is conventionally defined as the point where the filter function reaches half height. Generalizing Cohen's reparameterisation of the normal distribution to replace the standard deviation with a full-width half-maximum (FWHM) parameter \cite{cohen2019better}, we reparameterise the scale $\alpha$ with the FWHM parameter $h$. To derive the new form, we set the function value equal to 0.5 and solve for the solutions of $x$ (i.e. points/frequencies at which the function is at half height), depending on the center frequency $\mu$ and function shape $\beta$:
\begin{equation}
\begin{split}
    \frac{1}{2} = & e^{-(|x-\mu|/\alpha)^\beta} \\
    2^{-1} = & e^{-(|x-\mu|/\alpha)^\beta} \\
    - \ln(2) = & -(|x-\mu|/\alpha)^\beta \\
    - \ln(2) = & -\frac{|x-\mu|^\beta}{\alpha^\beta} \\
    \alpha^\beta\ln(2) = & |x-\mu|^\beta \\
    \alpha\ln(2)^{1/\beta} = & |x-\mu|,
\end{split}
\end{equation}
resulting in the two points $x_1$ and $x_2$
\begin{equation}
\begin{split}
    x_1 = & \mu - \alpha\ln(2)^{1/\beta} \\
    x_2 = & \mu + \alpha\ln(2)^{1/\beta}.
\end{split}
\end{equation}
The FWHM can then be defined as the distance between the two points at which the function reaches half height as
\begin{equation}
\label{eq:fwhm_reparameterization}
\begin{split}
    h = & x_2 - x_1 \\
    h = & \mu + \alpha\ln(2)^{1/\beta} - (\mu - \alpha\ln(2)^{1/\beta}) \\
    h = & 2\alpha\ln(2)^{1/\beta} \\
    \alpha = & \frac{h}{2\ln(2)^{1/\beta}}.
\end{split}
\end{equation}
This allows us to use the filter bandwidth to reparameterize the scale ($\alpha$) of the generalised Gaussian, which is more intuitive for signal processing applications. The filter $F$ is then finally defined as
\begin{equation}
\label{eq:generalized_gaussian_filter}
\begin{split}
    F(x) = & e^{-(|x-\mu|/\alpha)^\beta} \\
    \alpha = & \frac{h}{2\ln(2)^{1/\beta}}.
\end{split}
\end{equation}

At $\beta=2$ the generalised Gaussian equals the normal distribution, which corresponds to a Morlet wavelet in the time domain \cite{cohen2019better}. For higher values of the shape parameter, the filter moves towards a rectangular function, similar to a sinc filter in the frequency domain (see Fig. \ref{fig:filter_gradients}). In fact, the sinc filter is a special case of the generalized Gaussian in the limit of $\beta=\infty$. This can be shown by inserting the FWHM reparameterization based on the bandwidth $h$ (equation \ref{eq:fwhm_reparameterization}) for the scale parameter $\alpha$ into the generalised Gaussian function centered around $\mu$ as
\begin{equation}
\begin{split}
    F(x) = & e^{-(|x-\mu|/\alpha)^\beta} \\
    = & e^{-(|x-\mu|/\frac{h}{2\ln(2)^{1/\beta}})^\beta} \\
    = & e^{-\ln(2)(|x-\mu|/\frac{h}{2})^\beta}.
\end{split}
\end{equation}
It is then straightforward to analyze the function in the limit of $\beta\rightarrow\infty$:
\begin{equation}
\begin{split}
    \lim_{\beta\rightarrow\infty} e^{-\ln(2)(|x-\mu|/\frac{h}{2})^\beta} = &
    \begin{cases}
        1, & \textit{if $|x-\mu| < \frac{h}{2}$}; \\
        0.5, & \textit{if $|x-\mu| = \frac{h}{2}$}; \\
        0, & \textit{if $|x-\mu| > \frac{h}{2}$}.
    \end{cases}
\end{split}
\end{equation}
This precisely coincides with the frequency domain representation of an ideal sinc bandpass filter given by two heavy-side step functions centered around $\mu$ with bandwidth $h$.

We clamp the shape parameter at a minimum of $\beta=2$, since the generalised Gaussian moves towards the Laplace distribution for lower values, which is expected to interfere with training due to its non-continuous derivative.

The cost of a wide transition band is reduced frequency selectivity, which makes the function a less ideal filter. This potential problem can be addressed by making the shape parameter of the generalised Gaussian trainable, which can then move the filter from the Gaussian towards a rectangular shape over the course of training.

Designing the filter in the frequency domain also allows for full control over the phase response. For simplicity we use linear-phase filters here with a group delay of 20 ms, but a zero-phase or any linear or non-linear phase response could be used depending on the feature in question. The group delay of a linear-phase setup could also be added as a trainable parameter to allow for phase alignment between signals (not reported here).

We initialize all filters at 23 Hz with a wide bandwidth (44 Hz) and shape $\beta=2$, forming a normal distribution. The filter then has a non-zero derivative for all relevant frequencies. During training, filters narrow down to specific frequencies of interest. We accelerate the training of the shape parameter by rescaling it to
\begin{equation}
    \beta_{rescale} = 8\beta - 14.
\end{equation}
For numerical stability in the generalised Gaussian function, we clamp the shape parameter at a maximum of $\beta=3$ (corresponding to $\beta_{rescale}=10$).

The filters are trained independently for each electrode, except when using phase-locking value (PLV) connectivity, where each filter is shared across all electrodes to limit the model to within-frequency band connectivity as required by the definition of the metric. Whenever multiple filters per electrode  are used (and, hence, multiple feature maps are derived from the same multichannel signal, as shown in Fig. \ref{fig:EEGminer}), the feature module handles each feature map separately and in our correlation model inter-electrode connections are considered solely within individual maps. This results in multiple independent feature maps of connectivity, which are then concatenated into a single feature vector for classification.

\subsection{Differentiable Feature Module}

The proposed model is compatible with a variety of features, as long as they are defined for specific frequency bands and can be implemented in a differentiable way. For this study, we focus on band magnitude and signal correlations, and also perform initial tests on a phase synchrony descriptor, namely PLV.

The band-limited magnitude measure employed in this study is computed as the magnitude of the bandpass filtered signal, given by 
\begin{equation}
   M = \frac{1}{|\Omega|} \sum_\omega |x(\omega)|
\end{equation}
as the sum of magnitudes over frequency bins $\omega \in \Omega$, given the filtered signal $x$ in frequency domain. For simplicity, we average the magnitude over all frequency bins instead of just the bins within the limits of the bandpass filter. Using the magnitude instead of the power was observed to be more successful, which is likely due to the non-linearity of squaring, which also leads to skewed feature standardizations.

To assess functional connectivity between the filtered signals, a connectivity measure has to be chosen and implemented to be a well-behaved differentiable function in order to allow gradients to backpropagate through the measure into the trainable bandpass filters. A particularly suitable candidate are signal correlations due to their simplicity and relative success \cite{bakhshayesh2019detecting}, computed as the Pearson correlation between two bandpass filtered signals. Signal correlations are defined both as coupling within a frequency band as well as across different frequency bands, which makes them a natural fit to the independently trainable bandpass filters proposed here. The correlation function itself is well differentiable, as can be seen from the fact that it is even sometimes used as the loss function \cite{pandit2019many}. The correlation measure between two signals $X$ and $Y$ is given by
\begin{equation}
    r_{XY}
        = \frac{cov(X, Y)}{\sigma_X \sigma_Y}
        = \frac{\sum_t (x_t - \bar{x}) (y_t - \bar{y})}{\sqrt{\sum_t (x_t - \bar{x})^2} \sqrt{\sum_t (y_t - \bar{y})^2}}
\end{equation}
over time $t \in T$, where $\bar{x}$ and $\bar{y}$ are the signal means. This can be efficiently implemented using an inner product matrix multiplication for the numerator. Signal correlations are defined in the range [-1, 1], with larger absolute values indicating stronger coupling. Since a phase shift between two signals can lead to negative correlations, we take the absolute value ($|r_{XY}|$).

As an alternative estimate of functional connectivity, with a particular focus on the phase relationships between signals, we test phase locking values (PLV) \cite{lachaux1999measuring}. PLV is determined by the difference of the instantaneous phases $\Delta\phi$ of two signals as
\begin{equation}
    PLV = \frac{1}{T}|\sum_t e^{i\Delta\phi}|.
\end{equation}

Two signals with a strong coupling exhibit stable phase differences, leading to a high PLV, defined in the range [0, 1]. We build on the computationally efficient implementation proposed in \cite{bruna2018phase} (see also \cite{georgiadis2018exploiting}), while explicitly handling calculations with complex numbers as required by the PyTorch version used in this study. To decompose phase locking values (PLV) into efficient inner matrix products, we make use of the identity between the polar and Cartesian representation of the complex analytical signal, given by
\begin{equation}
    Ae^{i\phi} = x + iH(x),
\end{equation}
where $H(x)$ is the Hilbert transform and $A=\sqrt{x^2 + iH(x)^2}$ is the envelope of the analytical signal. We obtain the oscillatory component $e^{i\phi}$ by normalizing the analytical signal by the amplitude $A$ as
\begin{equation}
    e^{i\phi} = \frac{x + iH(x)}{A}
\end{equation}
and then define the real component $u$ and the imaginary component $iv$ of $e^{i\phi}$ as
\begin{equation}
\begin{split}
    u = & \frac{x}{A} \\
    iv = & i\frac{H(x)}{A}.
\end{split}
\end{equation}
The differences in instantaneous phase $\Delta\phi$ used in the PLV metric can then be expressed making use of the complex representation $u+iv$ as
\begin{equation}
\label{eq:plv}
\begin{split}
    P&LV = \frac{1}{T}|\sum_t e^{i\Delta\phi}| \\
    = & \frac{1}{T} |\sum_t e^{i\phi_1 - i\phi_2}| \\
    = & \frac{1}{T} |\sum_t e^{i\phi_1} e^{- i\phi_2}| \\
    = & \frac{1}{T} |\sum_t (u_1 + iv_1)(u_2 - iv_2)| \\
    = & \frac{1}{T} |\sum_t (u_{1}u_{2} - u_1iv_2 + u_2iv_1 - iv_1iv_2)| \\
    %= & \frac{1}{T} |\sum_t u_{1}u_{2} - \sum_t u_1iv_2 + \sum_t u_2iv_1 - \sum_t iv_1iv_2| \\
    = & \frac{1}{T} |(\sum_t u_{1}u_{2} + \sum_t v_1v_2) - i(\sum_t u_1v_2 - \sum_t u_2v_1)| \\
    %= & \frac{1}{T} \sqrt{(\sum_t u_{1}u_{2} + \sum_t v_1v_2)^2 + (\sum_t u_1v_2 - \sum_t u_2v_1)^2}.
\end{split}
\end{equation}
The resulting equation can be implemented based on four inner matrix products, handling the real and imaginary components separately.

% Volume conduction
At this point, it is important to notice that functional brain connectivity features, derived from EEG recordings in sensor space, are generally expected to be contaminated by volume conduction effects \cite{van1998volume}, which make signals recorded at nearby sensors to appear correlated even in the absence of actual neural connectivity \cite{srinivasan2007eeg}. Spurious connectivity, due to volume conduction, is expected to have only minimal impact on the features extracted via EEGminer for the following reasons. First, by training the classifiers to discriminate between two recording conditions (or groups), we extract only discriminative connectivity features and hence disregard spurious connectivity due to volume conduction, which is expected to behave like a constant/common factor \cite{cohen2014analyzing}. Aligned with this consideration, in our connectivity plots we present the mined edges weighted by classifier weights and not weighted by the absolute measure of connection strength. Second, by encompassing cross-frequency correlations in our correlation models, the influence of volume conduction is further reduced \cite{cohen2014analyzing}. Particularly for the results included in section \ref{sec:evaluation}, it is the spatial resolution of electrodes array in the case of MyBrainTunes dataset used for music vs rest and female vs male classification (see section \ref{sec:datasets}) and the cross-frequency character of the detected dependencies in SEED datasets that make dispensable the further signal manipulation for overcoming volume conduction. Additional steps that could be considered for mitigating the effects of volume conduction are appropriate preprocessing (e.g. by means of Laplacian transform \cite{nunez1994theoretical} or beamformers \cite{gross2001dynamic}) or modifying the differentiable feature module so as to overlook zero-lag correlations (e.g. by adopting the imaginary PLV \cite{bruna2018phase} version of eq. \ref{eq:plv}).

\subsection{Classifier and Feature Interpretation}

As a classifier we use a linear layer with sigmoid activation, essentially performing logistic regression. This has the advantage that the trained classifier weights can directly be used as importance weights attributed to the features. For non-linear classifiers the interpretation strategy would have to be adapted, e.g. by using prediction gradients with respect to the features \cite{lundberg2017unified}. Since different features can have different distributions, they need to be standardized in order to allow for the usage of the regression coefficients directly as an importance measure. In a machine learning setting with mini-batch gradient descent, standardization is performed via batch normalization \cite{ioffe2015batch}, which keeps a running mean and variance over mini-batches to approximate dataset statistics. In our case we use non-affine batch normalization, meaning the target mean and standard deviation are not trainable. We use a batch size of 256, since a larger batch size leads to a better estimation of the dataset statistics. To further improve model convergence and feature standardization, we perform cosine decay to zero on the learning rate \cite{loshchilov2016sgdr}. This ensures that the magnitude of model updates is reduced over time, allowing the batch normalization to capture final statistics of the discovered features.

For the band magnitude model we can investigate the full search space by plotting the relative difference in frequency magnitudes between the two classes. This allows us to check whether the discovered features are actually represented in the dataset. This is not possible for the connectivity model due to the vastly larger search space. We therefore save class-wise feature activations over the dataset and perform statistical t-tests to see whether the difference of feature means between the classes is statistically different from zero.

\subsection{Training Procedure}

The training of EEGminer models is performed via mini-batch stochastic gradient descent (SGD) with Nesterov momentum \cite{sutskever2013importance}. We found that heavy momentum on the bandpass filters helps training and improves performance. The momentum on the filters is set to 0.99, while it is set to the default value of 0.9 on the classifier. The increased momentum on the generalized Gaussian filters was beneficial for their convergence during training. The learning rate is initialized at 2e-3 and decayed to zero according to a cosine schedule. The learning rate decay is necessary for the batch normalization to get final estimates of statistical feature distributions and thereby provide standardized classification features. We train the EEGminer models for 5000 epochs on music vs rest, 300 on female vs male classification and 1200 epochs on positive vs negative emotion (these roughly correspond to the same number of total update steps). For the loss $\mathcal{L}$ we use the mean squared error (MSE) function between the targets $t$ and model predictions $x$ for simplicity. To sparsify discovered features, we use lasso regularization, which adds the L1 norm of the classifier weights $w$ to the loss with scaling factor $\gamma$, as defined by
\begin{equation}
    \mathcal{L} = \frac{1}{N}\sum_i(x_i - t_i)^2 + \gamma\sum_k |w_k|.
\end{equation}
The L1 penalty is tuned manually per model and classification task (EEGminer magnitude $\gamma$=2e-3 on all tasks, EEGminer signal correlations on music vs rest and female vs male classification $\gamma$=3e-3, on positive vs negative emotion $\gamma$=3e-2). The L1 penalty is mostly tuned for the desired sparsity of extracted features, while having only minor effects on classification performance. Note that no parameter regularization is performed on the bandpass filters.

\section{Experimental evaluation}
\label{sec:evaluation}

The usefulness of discovered features can be evaluated by the classification performance of EEGminer models on specific tasks. We benchmark the performance of the proposed models against conventional machine learning models and deep learning models that are commonly used in the EEG domain. This serves to illustrate whether the discovered features have similar discriminative power as conventional approaches with manual feature engineering as well as using modern black box classifiers. The aim is not to directly improve on decoding performance, but to provide a research tool for discovering features of brain activity that have discriminative power close to current state-of-the-art decoding models.

\subsection{Datasets}
\label{sec:datasets}

\begin{table*}[!t]
\renewcommand{\arraystretch}{1.3}
\centering
\caption{Validation performance (mean$\pm$std) of tested models on three different tasks. The best performing models on each task are highlighted with bold font. *The number of parameters for the SVM is given by the number of input features. **Due to the large standard deviation of results within folds (not given), results are here averaged over the last 10 epochs of training before computing mean and standard deviation across folds.}
\label{tab:validation_performance}
\begin{tabular}{lllllll}
\hline
                     & \multicolumn{2}{c}{Music vs Rest}  & \multicolumn{2}{c}{Female vs Male} & \multicolumn{2}{c}{Pos vs Neg Emotion} \\ \hline
                     & \# params & val UAR                & \# params & val UAR                & \# params   & val UAR                  \\ \hline
EEGminer magnitude   & 113       & 0.70$\pm$0.03          & 113       & 0.69$\pm$0.05          & 497         & \textbf{0.75$\pm$0.11}   \\
EEGminer correlation & 267       & 0.68$\pm$0.02          & 267       & 0.74$\pm$0.03          & 4,155       & 0.74$\pm$0.15            \\
EEGminer PLV         & 189       & 0.66$\pm$0.02          & 189       & 0.72$\pm$0.03          & 3,789       & \textbf{0.75$\pm$0.13}   \\ \hline
SVM magnitude        & 84*       & 0.68$\pm$0.04          & 84*       & 0.70$\pm$0.03          & 434*        & \textbf{0.75$\pm$0.12}   \\
SVM correlation      & 3,486*    & 0.67$\pm$0.03          & 3,486*    & 0.76$\pm$0.02          & 69,006*     & 0.73$\pm$0.16            \\ \hline
ShallowConvNet       & 42,001    & \textbf{0.71}$\pm$0.05 & 42,001    & 0.74$\pm$0.03          & 112,401     & 0.71$\pm$0.09**          \\
EEGNet               & 1,573     & 0.67$\pm$0.05          & 3,273     & \textbf{0.77$\pm$0.02} & 1,637       & 0.67$\pm$0.14**          \\
SyncNet              & 1,261     & 0.66$\pm$0.04          & 1,261     & 0.71$\pm$0.02          & 1,901       & 0.71$\pm$0.09**          \\
SPDNet               & 785       & 0.66$\pm$0.03          & 785       & 0.69$\pm$0.04          & 1,761       & 0.68$\pm$0.12            \\ \hline
\end{tabular}
\end{table*}

%[MyBrainTunes dataset]
The dataset introduced here was formed from the data recorded during a very recent large-scale experiment, organized by two of the authors of this study and collected in cooperation with the Science Museum London\footnote{https://blog.sciencemuseum.org.uk/music-and-the-brain/}. The experiment was approved by the Science, Engineering and Technology Research Ethics Committee (SETREC) at Imperial College London. It consists of brain activity recorded from 761 volunteer subjects (400 female, 361 male) aged 12 and above (mean=28.7, std=11.0). The experiment followed a passive music-listening design, similar to \cite{bakas2021estimate}\cite{adamos2016towards}. %Each subject first listened to 30 second segments of 30 songs, indicated their familiarity with the song and gave a rating of 0 for disliked songs or a rating from 1 to 5 for the degree of liking.
A 30 second EEG baseline was recorded for each subject, after which 30 second segments of 30 songs were played in random order while recording EEG activity, for a total of 30x30=900 seconds of music listening EEG per subject. During EEG recording, participants were asked to remain still to reduce recording artifacts, to keep their eyes open and to focus on a dot in the middle of a screen to avoid introducing noise from the muscular activity of eye movements. The recordings were conducted with Emotiv EPOC+ wet electrode headsets with 14 channels at 256 Hz (AF3, F7, F3, FC5, T7, P7, O1, O2, P8, T8, FC6, F4, F8, AF4 in accordance with the 10-20 system). %In addition to the music ratings, participants were asked to fill out a questionnaire with basic information including their age, sex, handedness, ethnicity, educational level, musical experience and coffee intake in the last few hours.
The song playlist was pre-defined and comprised of songs from the top of the UK charts of the last decades including songs from various genres (e.g. pop, rock, ballads, hip-hop, etc.).

The female vs male classification task uses all recordings of the MyBrainTunes dataset. For the music vs rest task a single song recording is randomly chosen per subject to provide an equal number of song and resting state trials.

%[SEED dataset]
The SEED dataset contains EEG recordings from 15 participants while watching emotional film clips \cite{duan2013differential}\cite{zheng2015investigating}. Each subject participated in 3 recording sessions, viewing the same 15 clips with negative, neutral or positive emotional content. The EEG data provided include recordings from 62 electrodes at 200 Hz sampling rate. The power line noise at 50 Hz has been removed. For the positive vs negative emotion task we discard clips with neutral emotion. We break up the long recordings into windows of 20 seconds. Windows are sampled randomly and treated independently for training samples. To account for this windowing, every recording is sampled 9 times per epoch, which increases the total number of update steps. For validation samples, each recording is split into consecutive non-overlapping windows and the model predictions are averaged across these windows.

%[Preprocessing]
Both datasets have been preprocessed as follows: the multichannel signals were filtered (1-45 Hz for our dataset, 0.5-99.5 Hz for SEED) using a zero-phase band-pass filter (3rd order Butterworth). To remove artifacts we resorted to independent component analysis (ICA) \cite{delorme2007enhanced}. Artifact suppression was carried out separately for each trial, based on an in-house implementation of wavelet-enhanced ICA (wICA \cite{castellanos2006recovering}). Specifically, independent components (ICs) were extracted from the multichannel signal by means of the EEGLAB Matlab toolbox\footnote{https://sccn.ucsd.edu/eeglab/}. Subsequently, wavelet decomposition based on wavelets of the biorthogonal family and wavelet shrinkage with a hard threshold based on false discovery rate \cite{abramovich1996adaptive} were applied to each one of the ICs. The multichannel signal was then reconstructed based on the artifact-free ICs. Hence, in this study, the use of “wICA-cleaned” brain activity is implied. The EEG signals were re-referenced by common average referencing. We downsample the MyBrainTunes dataset to 128 Hz. For the deep learning baseline models, the SEED dataset was also resampled to 128 Hz. For all models, inputs are standardized by a total trial mean and standard deviation. For SPDNet, the input is bandpass filtered in the 4-40Hz range before computing the spatial covariance matrix.

\subsection{Evaluation Procedure}

To equalize small class imbalances during training, we perform oversampling of underrepresented classes. Cross-validation was performed for all models, using 10 random folds with held-out subjects for the respective validation set. This means that for each fold, none of the samples of the subjects in the corresponding validation set were seen during training (also considering oversampling and the three sessions per subject on SEED), and results should be seen as subject-independent. The main performance metric is given as the unweighted average recall (UAR) at the end of training, reporting the mean and standard deviation across folds. The UAR for classes $c \in C, |C|=N$ is given by
\begin{equation}
    UAR = \frac{1}{N} \sum_c \frac{true\_positives_c}{true\_positives_c + false\_negatives_c}.
\end{equation}

\begin{figure*}[!t]
    \centering
    \includegraphics[width=0.9\textwidth]{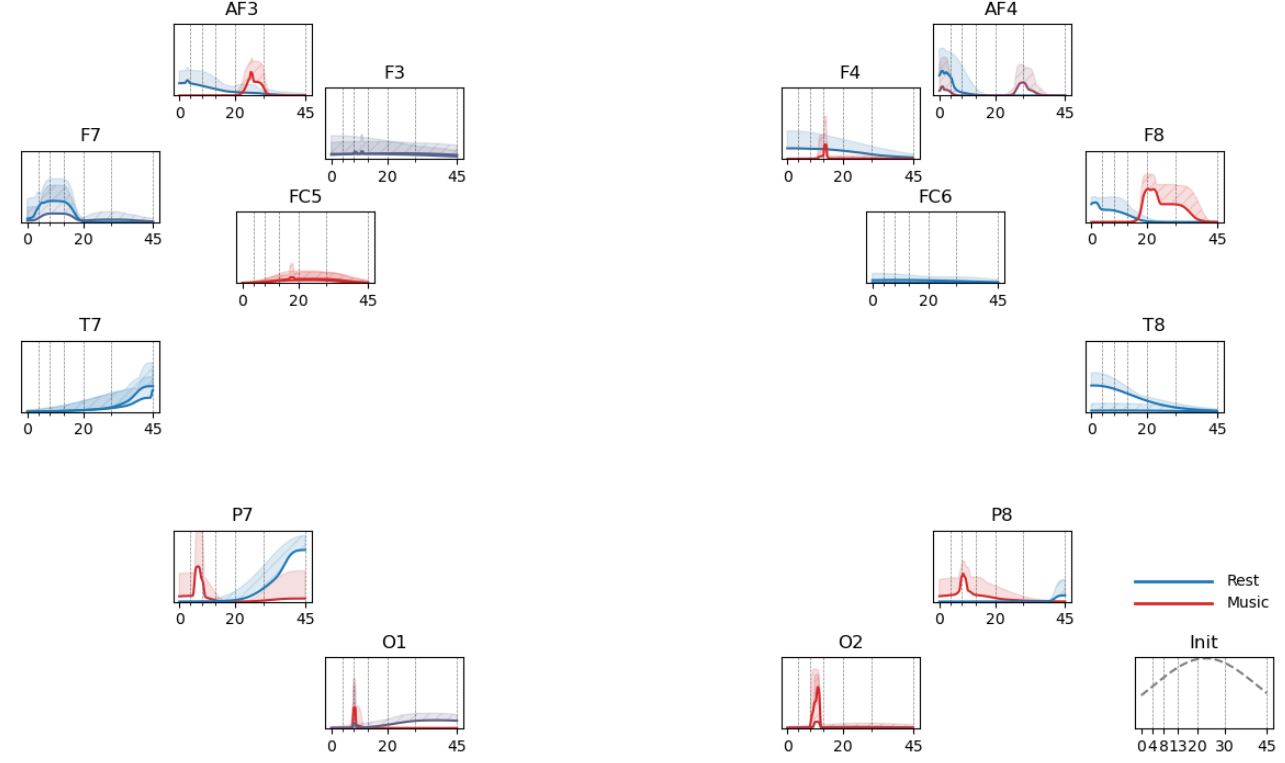}
    \caption{Trained magnitude EEGminer model on music listening vs resting state classification, indicating important features of brain activity. Filter lines show the averaged filters across folds scaled by the model weights, with shaded areas denoting the maximum weight and colors indicating the feature direction. Since two freely moving filters were used per electrode, averaging of the filters over folds is performed according to the respective center frequency (i.e. average all low-center filters and all high-center filters). The inset shows the filter initialization. The limits of canonical EEG bands are given as dotted lines.}
    \label{fig:1244_music_power_contributors}
\end{figure*}

\subsection{SVM and Deep Learning Baselines}

%[SVM]
As a performance baseline from the conventional neuroscience perspective, we employ a linear support vector machine (SVM) on standardized features spanning the whole search space. Regarding band magnitude, the features are the signal magnitudes of all channels filtered in canonical EEG bands\footnote{$\delta$ 1-4 Hz, $\theta$ 4-8 Hz, $\alpha$ 8-13 Hz, $\beta_1$ 13-20 Hz, $\beta_2$ 20-30 Hz, $\gamma$ $>$30 Hz}. In the signal correlation case, the features are all possible connections across all sensors filtered within and across all EEG bands (3,486 for our dataset, 69,006 for SEED). Performance was evaluated in the same way as for the EEGminer and deep learning models, using 10-fold subject-independent cross-validation.

\begin{figure*}[!t]
    \centering
    \includegraphics[width=0.9\textwidth]{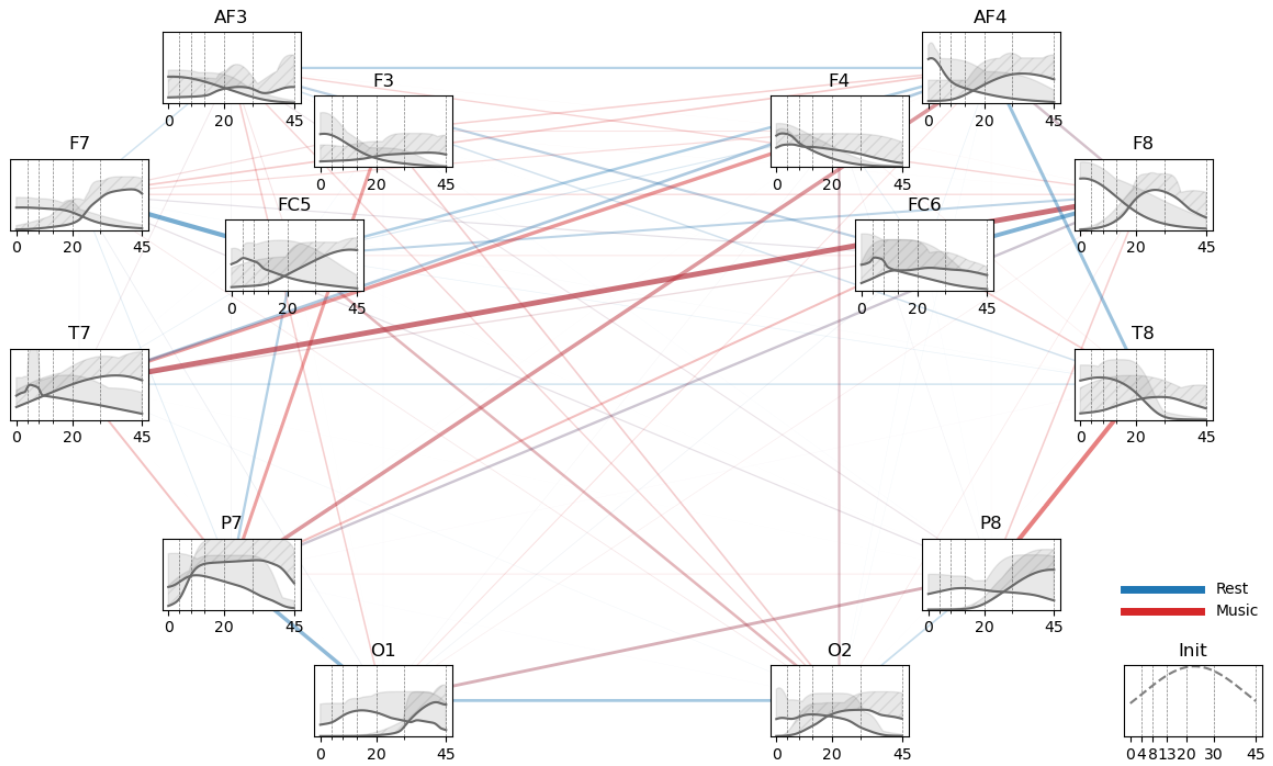}
    \caption{Trained correlation EEGminer model on music listening vs resting state classification, indicating important features of brain activity. Connections are scaled by the mean model weights across folds, with the colors indicating the feature direction. Filter lines show the averaged filters across folds scaled by the model weights on the strongest adjacent connection, with shaded areas denoting the maximum weight. Since two freely moving filters were used per electrode, averaging of the filters over folds is performed according to the respective center frequency (i.e. average all low-center filters and all high-center filters), and the stronger of the two model weights is used for the connection. The inset shows the filter initialization. The limits of canonical EEG bands are given as dotted lines.}
    \label{fig:1255_music_sigcorr_contributors}
\end{figure*}

%[Deep learning benchmarks]
To benchmark our approach against deep learning models, we train ShallowConvNet, EEGNet, SyncNet and SPDNet on all three tasks. DeepConvNet is omitted from the baselines here due to inferior performance when compared to ShallowConvNet on the studied tasks. We adjust the filter kernel sizes of ShallowConvNet and EEGNet for input signals sampled at 128Hz. All four deep learning models are trained for 100 epochs using an Adam optimizer. In the case of the SPDNet, an Adam optimizer with manifold constraint was used\footnote{https://geoopt.readthedocs.io/en/latest/optimizers.html}. The optimizer learning rate is 1e-3 for ShallowConvNet and EEGNet and 2e-3 for SyncNet, as proposed by the respective authors. The learning rate used for SPDNet is 1e-3. The loss function is binary cross-entropy between targets and predictions for all deep learning models. ShallowConvNet and EEGNet use a batch size of 16, while SyncNet uses 10 and SPDNet uses 30. The dropout rates for ShallowConvNet and EEGNet are 0.5 and 0.25 respectively. SyncNet uses dropout of input channels, which we employ with a rate of 0.25 only on the SEED dataset, as it gave inferior performance when used on the MyBrainTunes dataset (likely due to the small number of electrodes). No weight decay is used. For EEGNet we use 4 temporal filters on music vs rest and positive vs negative emotion as the number of training samples is somewhat limited, while we use 8 temporal filters on female vs male classification. For SPDNet we use 3 BiMap layers, with hidden size of 14 on the MyBrainTunes tasks and 16 on SEED.

\subsection{Performance Evaluation}

%[EEGminer vs SVM]
We evaluate the performance of all tested models using cross-validation and the unweighted average recall (UAR) metric. Table \ref{tab:validation_performance} illustrates the cross-validation performance on all three tasks. The performance of the proposed EEGminer models is on par with a linear support vector machine (SVM) \cite{cortes1995support} trained on pre-computed signal magnitude or signal correlations using canonical EEG bands. This shows that EEGminer models can discover the relevant features to match the performance of a conventional machine learning model that operates on a large pre-computed search space, while having a focus on clear feature interpretability.

%[EEGminer vs deep learning]
Furthermore, the classification performance of EEGminer models is on a comparable level to the deep learning baselines, performing marginally better or worse depending on the given task. The temporal resolution that convolutional neural networks offer is well-suited for stimulus-locked recordings, such as event-related potentials (ERPs) or motor imagery (MI), but might have relatively limited use on the spontaneous nature of the classification tasks studied here. This indicates that our proposed model limits trainable components to the relevant feature space, cutting unnecessary capacity and thereby reducing the potential for overfitting.

%[Correlations vs PLV]
Regarding connectivity features, both signal correlations and PLV achieved similar performance, with signal correlations having a small edge. This disparity could be due to the fact that PLV is limited to within-band connectivity, while signal correlations are also defined across frequencies and therefore have a larger feature space. We therefore focus on band magnitudes and signal correlations when discussing the mined features.

\subsection{Mined Features}

\begin{figure*}[!t]
    \centering
    \includegraphics[width=0.9\textwidth]{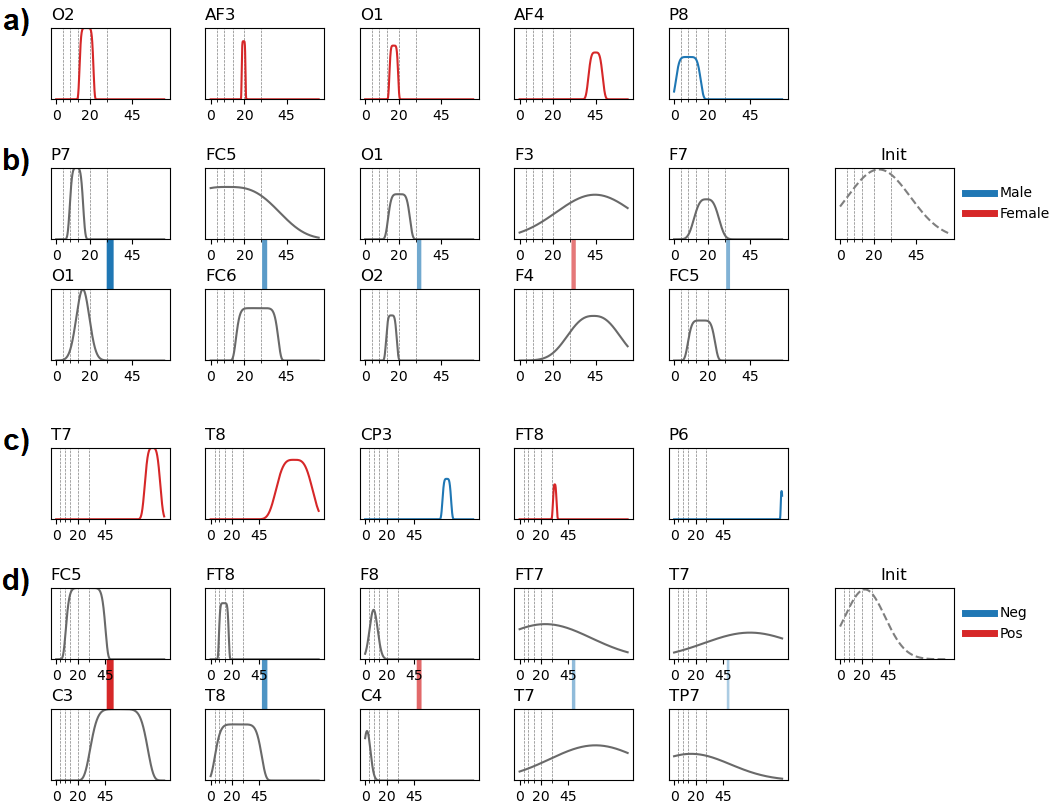}
    \caption{The five most important features discovered by the EEGminer models. The height of filter lines and scale of connections respectively indicate the model weight for each feature, with colors indicating the feature direction. Panel a) shows the magnitude EEGminer features on female vs male classification during music listening and b) shows the respective correlation EEGminer features. Panel c) shows the magnitude EEGminer features on positive vs negative emotion classification and d) shows the respective correlation EEGminer features. The insets show the respective filter initializations. The limits of canonical EEG bands are given as dotted lines.}
    \label{fig:top_features_MF_emotion}
\end{figure*}

%[Introduction]
Benefiting from the clear interpretability of EEGminer models, we can investigate the trained features in terms of relevant frequencies and functional connectivity. On the music vs rest classification task we present the full feature space as utilized by the respective models, while we focus on the most impactful features for female vs male classification during music listening and positive vs negative emotion recognition. When plotting top features on male vs female and positive vs negative emotion classification, averaging across folds is difficult. We reverted to the fold with highest validation performance. Using two filter channels can result in duplicate features, which we manually removed.

%[General trainability observations]
Looking at the trained magnitude EEGminer model on music vs rest classification (Fig. \ref{fig:1244_music_power_contributors}), it is apparent that the proposed generalised Gaussian filters exhibit strong trainability of the center frequency, bandwidth and shape, departing substantially from their initialized state. Many of the filters with high model weight narrow down to specific EEG bands (e.g. alpha in O2, theta in P7 and high beta in AF3). Multiple bands of interest can be found for the same sensor when using a model with multiple filter channels (e.g. P7, F8). Features are sparsified by L1 regularization on the linear classifier, forcing the model to disregard locations with minor contribution (e.g. FC5, FC6 and F3). The trained filters also exhibit good stability over folds, as seen in the small difference of mean and max filter importance, meaning the discovered features are consistent across reruns on other training folds.

%[Magnitude features music vs rest]
Relevant band magnitude features as discovered by the EEGminer model on music vs rest classification are decreased activity of lower frequencies in frontal electrodes (AF3, AF4, F7 and F8) during music listening and increased activity in high beta and gamma (AF3, AF4 and F8). Also used by the model are increased alpha activity in occipital electrodes (O1 and O2) during music listening and decreased gamma as well as increased theta activity in P7. Decreased gamma activity in T7 and in low frequencies in T8 are also associated with music listening.

\begin{figure*}[!t]
    \centering
    \includegraphics[width=\textwidth]{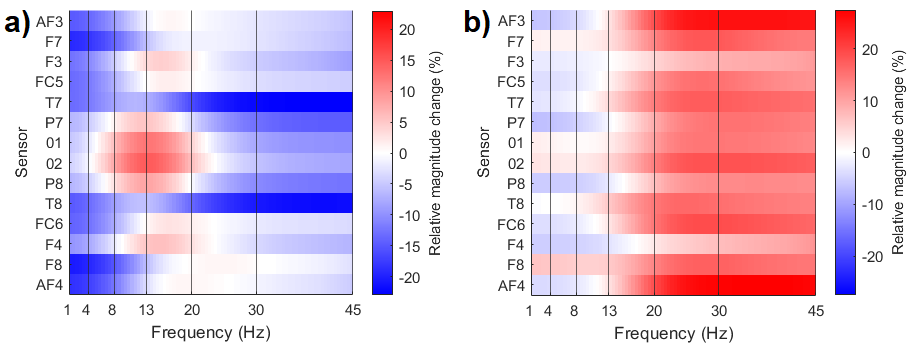}
    \caption{Relative change in grand averaged magnitude profiles for a) music-listening relative to resting state and b) female relative to male subjects during music listening.}
    \label{fig:mag_profiles_music_rest_MF}
\end{figure*}

%[Correlation features music vs rest]
When using signal correlations as the feature module to classify music listening vs resting state (Fig. \ref{fig:1255_music_sigcorr_contributors}), the most relevant features are cross-hemispheric connections between the left temporal/parietal with the right frontal areas. Most specifically during music listening there is increased coupling between high-frequency activity in T7 and low frequencies in F8. Also relevant is decreased local connectivity of gamma brainwaves between FC5 and F7 and of high frequencies in FC6 with low frequencies in F8. It can be observed that the model identifies a hemispheric specialisation of the temporal areas, focusing on cross-hemispheric connections of the left temporal area and within-hemisphere connections of the right temporal area. This agrees with the previously proposed respective specialisation of the temporal lobes regarding music perception \cite{zatorre2002structure}.

%[Features MF classification]
Regarding female vs male classification during music listening, the magnitude EEGminer model focuses on increased low beta activity in occipital areas (O1 and O2) for female subjects, as well as increased high frequencies in anterior frontal electrodes (beta in AF3 and gamma in AF4; see Fig. \ref{fig:top_features_MF_emotion}). Increased activity of higher frequencies in female subjects has previously been reported in studies of resting state EEG \cite{van2018predicting}. The model also finds decreased theta and alpha activity for female subjects in P8. On the other hand, the correlation EEGminer model mostly focuses on decreased connectivity in lower frequencies for females and increased connectivity in the gamma band, particularly in cross-hemispheric symmetric connections. Specifically, reduced connectivity between alpha in P7 and low beta in O1 is associated with female subjects, as well as reduced connectivity between FC5 and FC6 and reduced beta connectivity between O1 and O2 and FC5 and F7. Increased connectivity in female subjects is found in gamma activity between F3 and F4.

%[Features positive vs negative emotion]
The magnitude EEGminer model uses differences in gamma activity to classify emotion (see Fig. \ref{fig:top_features_MF_emotion}). Positive emotion is associated with increased gamma activity in temporal areas (T7, T8 and FT8) compared to negative emotion, as well as decreased gamma activity in parietal electrodes (CP3 and P6). The association of emotional state and high frequency activity in the temporal areas has previously been reported on the SEED dataset \cite{zheng2017identifying}\cite{li2018exploring}. The most important couplings detected by the correlation EEGminer model are increased within-hemisphere connectivity during positive emotion between central and frontal areas (C3 with FC5 and C4 with F8) and decreased within-hemisphere connectivity of temporal with adjacent electrodes (FT8 with T8, FT7 with T7 and TP7 with T7).

\subsection{Feature Validation}

\begin{figure*}[!t]
    \centering
    \includegraphics[width=\textwidth]{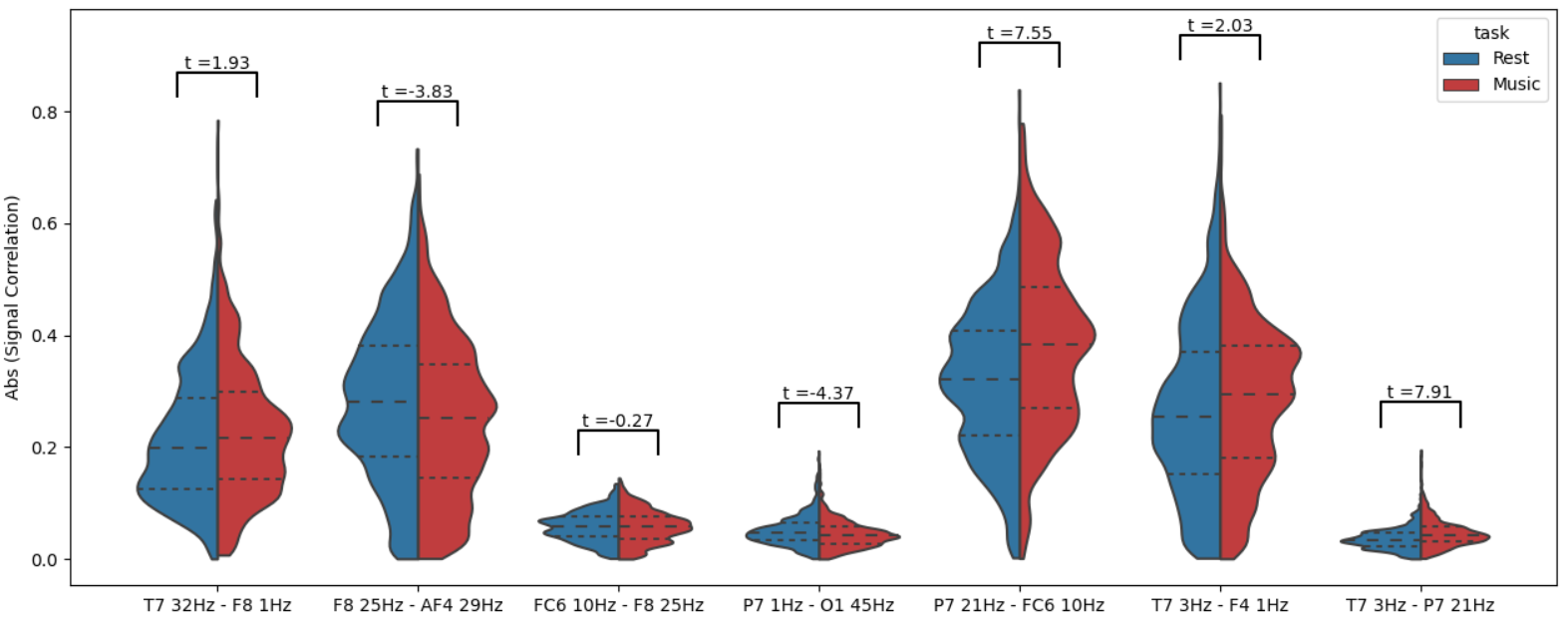}
    \caption{Violin plots of unstandardised top features discovered by the correlation EEGminer model trained on music vs rest classification. Dashed lines indicate the quartiles of each feature distribution on all samples, including train and test set. The associated statistical difference between classes is given by an independent two-samples t-test above each distribution.}
    \label{fig:1255_music_sigcorr_ttests}
\end{figure*}

\begin{figure*}[!t]
    \centering
    \includegraphics[width=\textwidth]{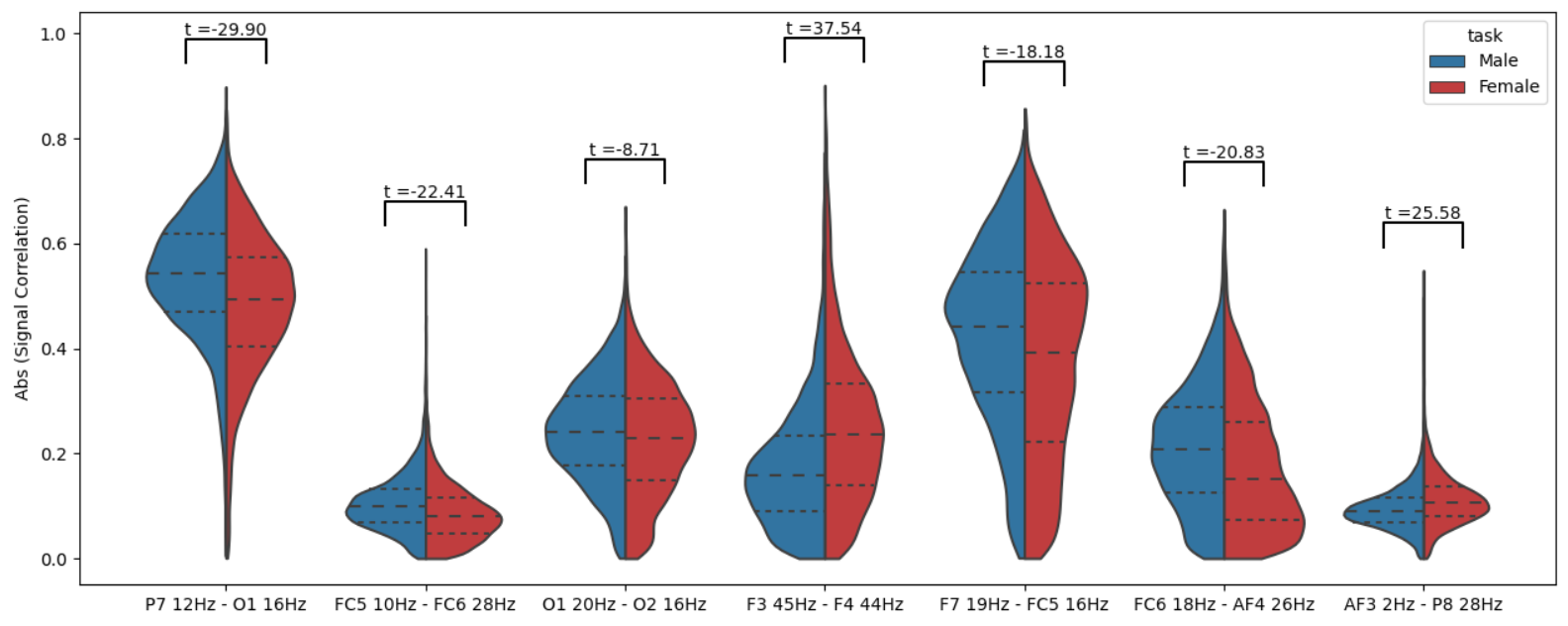}
    \caption{Violin plots of unstandardised top features discovered by the correlation EEGminer model trained on female vs male classification during music listening. Dashed lines indicate the quartiles of each feature distribution on all samples, including train and test set. The associated statistical difference between classes is given by an independent two-samples t-test above each distribution.}
    \label{fig:1258_MF_sigcorr_ttests}
\end{figure*}

To validate the features found by EEGminer models on our dataset (music vs rest, female vs male classification during music listening), we created the relative change of grand averaged magnitude profiles for the magnitude models, and violin plots with t-test statistics showing the distributions of the features used by signal correlation models. Since the search space for connectivity models is exceedingly large, we limit ourselves to the top features here.

Fig. \ref{fig:mag_profiles_music_rest_MF} (a) illustrates the relative change in grand averaged magnitude profiles during music listening relative to resting states for all sensors and available frequencies. Most of the features mined by the model are strongly represented in the grand average (reduced gamma in P7, increased alpha in O1 and O2, decreased delta to alpha in F7 and F8 and increased high beta and gamma in F8). Some mined features are ambiguous in the grand average, specifically increased theta in P7, which is shown in the alpha band by the grand average, and increased high beta in AF4, which is shown in low beta by the grand average.

The relative change in grand averaged magnitude profiles for female relative to male subjects during music listening is given in Fig. \ref{fig:mag_profiles_music_rest_MF} (b). All five of the top features discovered by the EEGminer model are strongly represented by the differences in grand averages (increased low beta in O1 and O2 for females, increased beta in AF3 and gamma in AF4, reduced theta and alpha in P8).

Violin plots showing the distributions of music listening and resting state EEG across all subjects (rest and one song each) of top features discovered by the EEGminer signal correlation model are given in Fig.  \ref{fig:1255_music_sigcorr_ttests}. All feature directions agree between the EEGminer model and the feature distributions in the dataset, although some of the features show low statistical significance in the difference of means. This can be attributed to the limited number of training examples using only one randomly chosen song per subject to equalize class balance, as well as the somewhat limited classification performance of the model (0.68$\pm$0.02).

Fig. \ref{fig:1258_MF_sigcorr_ttests} shows the distributions of top features discovered by the EEGminer signal correlation model for male and female subjects during music listening. All feature directions agree between the model and the feature distributions and all of the top seven features show strong statistical mean differences in the t-tests.

\section{Conclusion}
\label{sec:conclusion}

%[Summary and future]
We presented trainable filters parameterized by the generalised Gaussian function, which include Morlet wavelets and sinc filters as special cases and allow the model to select the appropriate filter during end-to-end training. Using differentiable implementations of neuroscientifically plausible EEG features like band magnitude and signal correlations, our proposed model can discover classification-relevant frequency bands and functional connectivity patterns among a large repertoire of possible features, matching the decoding accuracy of established deep learning models. The trained classifier further offers clear insights into feature importance and interpretations. In future work, going beyond the static nature of features used here, a feature mining mechanism with higher time resolution should be developed for event-related tasks.

%ConnEEGminer in front of SPDNet
The current application of deep learning models on SPD matrices, like SPDNet, in the context of EEG is limited by the choice of input matrix. In the context of this paper, a wide-band covariance matrix was used as a reasonable choice. This approach could benefit significantly from the trainable filters and connectivity estimation developed for EEGminer, which in the case of signal correlations produces SPD matrices. This exploration is left for future work.

%[Relevance]
Implementing an EEG pipeline as a differentiable deep learning model could bridge the gap between conventional research in neuroscience and difficult to interpret black-box methods commonly employed in the deep learning field. Such a model promises comparable performance with deep learning models as well as neuroscientific insights. This has important implications in practical applications as well, where model interpretability is often a strong requirement.

\ifCLASSOPTIONcompsoc
  \section*{Acknowledgments}
\else
  \section*{Acknowledgment}
\fi
We would like to thank the Science Museum London for hosting the data collection experiment.

\begin{comment}
\renewcommand{\abstractname}{Author Contribution}
\begin{abstract}
    These authors contributed equally: S. L. and S. B. Experiment design: D. A. and S. Z. Supervision: D. A., S. Z. and N. L. Data preprocessing: S. B., S. L., D. A. and N. L. Methodology: S. L., S. B. and Y. P. Interpretations: S. B., S. L, N. L. and D. A. Writing original draft: S. L. and S. B. Writing revisions: S. L., S. B., N. L. and Y. P.
\end{abstract}

\renewcommand{\abstractname}{Competing Interests Statement}
\begin{abstract}
    The authors declare no competing interests.
\end{abstract}
\end{comment}

\ifCLASSOPTIONcaptionsoff
  \newpage
\fi

% break references at x to balance columns
%\IEEEtriggeratref{49}

\bibliographystyle{IEEEtran}
\bibliography{IEEEabrv, references}

%Include IEEE membership

\begin{IEEEbiography}[{\includegraphics[width=1in,height=1.25in,clip,keepaspectratio]{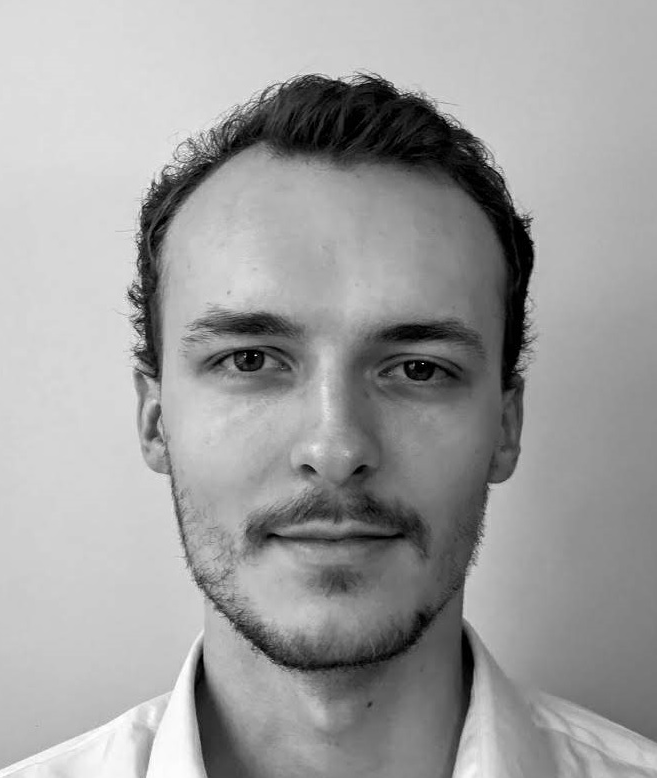}}]%
{Siegfried Ludwig} obtained a B.Sc. in International Business Administration and Entrepreneurship from Leuphana University Lueneburg, Germany, in 2018, and a M.Sc. in Artificial Intelligence from Radboud University Nijmegen, Netherlands, in 2020. Since 2021 he is a Ph.D. student at the Department of Computing, Imperial College London, U.K. His main interest is currently in developing novel deep learning methodologies for EEG decoding, with previous conference publications in natural language processing and neuromorphic computing.
\end{IEEEbiography}

\begin{IEEEbiography}[{\includegraphics[width=1in,height=1.25in,clip,keepaspectratio]{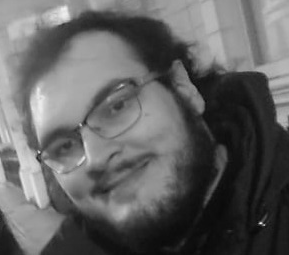}}]%
{Stylianos Bakas} received his M.Eng. in Electrical and Computer Engineering from the Aristotle University of Thessaloniki in 2016, and his M.Sc. in Digital Media - Computational Intelligence from the School of Informatics of the Aristotle University of Thessaloniki in 2020. He is currently a Ph.D. student at the School of Informatics of the Aristotle University of Thessaloniki. His research interests include developing novel signal processing and deep learning methodologies for applications in Brain-Computer Interfaces, brain connectomics and music neuroscience.
\end{IEEEbiography}

\begin{IEEEbiography}[{\includegraphics[width=1in,height=1.25in,clip,keepaspectratio]{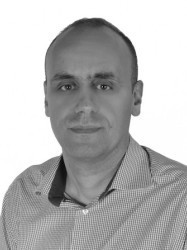}}]%(Member, IEEE)
{Dimitrios A. Adamos} is an Honorary Research Fellow for the Department of Computing of Imperial College London and leader of the $\#$\textit{MyBrainTunes} experiment held in London’s Science Museum. He is also a Senior Research fellow at the School of Music Studies, Aristotle University of Thessaloniki (AUTh), Greece. He holds a M.Eng. in Electrical \& Computer Engineering, an M.Sc. in Medical Informatics from the School of Medicine and a Ph.D. in Neuroinformatics from the School of Biology of AUTh. His current research includes deep learning and brain connectomics.
\end{IEEEbiography}

%\vfill
%\newpage

\begin{IEEEbiography}[{\includegraphics[width=1in,height=1.25in,clip,keepaspectratio]{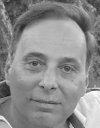}}]%
{Nikolaos Laskaris} is currently an Associate Professor with the Department of Informatics, Aristotle University, Greece. He is a member of AIIA Laboratory and leads the Neuroinformatics.GRoup. He is a coauthor of more than 100 scientific publications. His current research interests include neuroinformatics, brain connectomics and the applications of machine learning, data mining and nonlinear dynamics in biomedicine and neuroscience.
\end{IEEEbiography}

\begin{IEEEbiography}[{\includegraphics[width=1in,height=1.25in,clip,keepaspectratio]{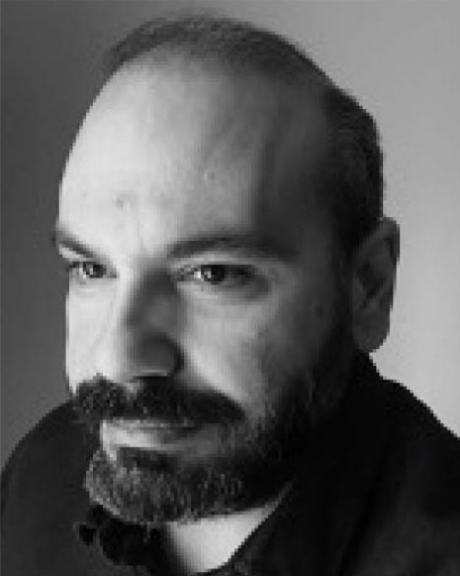}}]%(Member, IEEE)
{Yannis Panagakis} received the B.Sc. degree in informatics and telecommunications from the University of Athens, Greece, and the M.Sc. and Ph.D. degrees in computer science from the Aristotle University of Thessaloniki, Greece. He previously held research and academic positions at the Samsung AI Centre, Cambridge, U.K., Middlesex University London, and Imperial College London, U.K. He is currently an Associate Professor of machine learning and signal processing with the National and Kapodistrian University of Athens. His research focuses on machine learning and its interface with tensor methods, signal processing, and optimization. He has published over 80 articles in leading journals and conferences.
\end{IEEEbiography}

\begin{IEEEbiography}[{\includegraphics[width=1in,height=1.25in,clip,keepaspectratio]{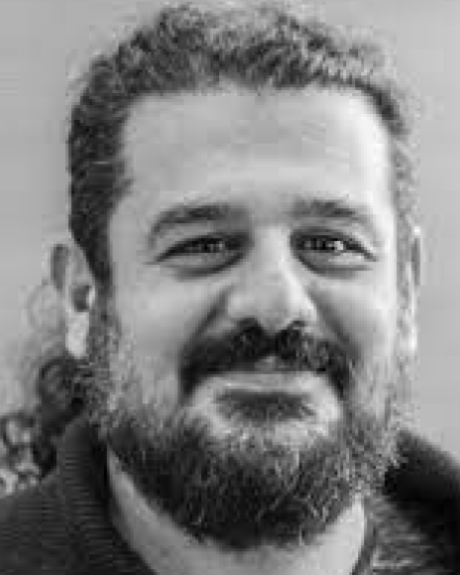}}]%(Member, IEEE)
{Stefanos Zafeiriou} was also a Distinguishing Research Fellow with the University of Oulu, Oulu, Finland, under the Finish Distinguishing Professor Programme from 2016 to 2020. He was the cofounder of two startups Facesoft and Ariel AI (exited to Snap). He is currently a Professor of machine learning and computer vision with the Department of Computing, Imperial College London, London, U.K., and an EPSRC Early Career Research Fellow. He has more than 15k citations to his work, with an H-index of 59. He has coauthored more than 80 journal articles mainly on novel statistical machine learning methodologies applied to computer vision problems, such as 2-D/3-D face analysis, deformable object fitting and tracking, shape from shading, and human behavior analysis, published in the most prestigious journals in his fields of research, such as the IEEE T-PAMI, the International Journal of Computer Vision, and many papers in top conferences, such as Conference on Computer Vision and Pattern Recognition (CVPR), International Conference on Computer Vision (ICCV), European Conference on Computer Vision (ECCV), and International Conference on Machine Learning (ICML).

Dr. Zafeiriou was a recipient of the Prestigious Junior Research Fellowships from Imperial College London in 2011 and the President’s Medal for Excellence in Research Supervision for 2016. His students are frequent recipients of very prestigious and highly competitive fellowships, such as the Google Fellowship x2, the Intel Fellowship, and the Qualcomm Fellowship x4. He was the General Chair of the British Machine Vision Conference (BMVC) in 2017. He has served an Associate Editor and a Guest Editor in various journals, including IEEE TRANSACTIONS ON PATTERN ANALYSIS AND MACHINE INTELLIGENCE, International Journal of Computer Vision, IEEE TRANSACTIONS ON AFFECTIVE COMPUTING, Computer Vision and Image Understanding, IEEE TRANSACTIONS ON CYBERNETICS, the Image and Vision Computing journal. He has been a Guest Editor of more than eight journal special issues and co-organized over 16 workshops/special sessions on specialized computer vision topics in top venues, such as CVPR/IEEE International Conference on Automatic Face \& Gesture Recognition (FG)/ICCV/ECCV (including three very successfully challenges run in ICCV’13, ICCV’15, and CVPR’17 on facial landmark localization/tracking).
\end{IEEEbiography}

%\vfill

%\renewcommand{\figurename}{Supplementary Figure}
%\setcounter{figure}{0}

%\newpage

%\appendix

\end{document}